\documentclass[11pt]{article}
\usepackage{graphicx}
\usepackage{natbib}
\usepackage{palatino}

\newcommand{\ignore}[1]{}

\usepackage{fancyhdr}
\begin{document}

\noindent {\Large Technical Report UTEP-CS-24-23}

\bigskip\bigskip\medskip

\noindent {\huge \bf Topics in the Study of the Pragmatic Functions
  of}

\medskip 
\noindent {\huge \bf  Phonetic Reduction  in Dialog}

\bigskip\bigskip \noindent  Nigel G. Ward, Carlos A. Ortega

\medskip

\noindent Department of Computer Science, University of Texas at El Paso 

\medskip
\noindent{nigelward@acm.org,  carlos.ortega2001@hotmail.com}

\noindent  \today

\noindent \bigskip 

\noindent {\bf Abstract}

\smallskip \noindent Reduced articulatory precision is common in
speech, but for dialog its acoustic properties and pragmatic functions
have been little studied.  We here try to remedy this gap.  This technical
report contains content that was omitted from the journal article
\citep{reduction-article}.  Specifically, we here report 1) lessons
learned about annotating for perceived reduction, 2) the finding that,
unlike in read speech, the correlates of reduction in dialog include
high pitch, wide pitch range, and intensity, and 3) a baseline model
for predicting reduction in dialog, using simple acoustic/prosodic
features, that achieves correlations with human perceptions of 0.24
for English, and 0.17 for Spanish.  We also provide examples of
additional possible pragmatic functions of reduction in English, and
various discussion, observations and speculations.

\smallskip \noindent {\bf Keywords:} reduced articulatory precision,
hypoarticulation, prosody, pragmatic functions, corpus study,
annotation, perceptions, English, Spanish, correlations, predictive
model

\smallskip

\bigskip\noindent
{\bf Contents}
\smallskip

\begin{tabular}{ll}
1 & Introduction \\
2 & Related Research \\
3 & Annotating for Perceived Reduction \\
4 & Features and Correlations \\
5 & Predictive Models \\
6 & More on the Functional Annotation Process \\
7 & More on the Pragmatic Functions of Reduction in English \\
8 & More on the Experiment \\
9 & Reduction and Negative Assessment  in Spanish \\
10 &  Summary  
\end{tabular}

\section{Introduction} 

Feeling that our inventory of prosodic features was incomplete, we set
out to add phonetic reduction to the features handled by the Midlevel
Prosodic Features Toolkit \citep{midlevel-toolkit23}.  We failed in
this goal, but in the process learned a lot about reduction.  The
headline finding was the result that phonetic reduction correlates
with positive assessments in American English, and that result, plus
closely related topics, reported in a journal article submission
\citep{reduction-article}.  However, not everything that we learned
fit there, however, so this document reports the rest.  Some of the
discussions are stand-alone --- notably those of spectral tilt,
annotation for reduction, and prosodic correlates of reduction, as
found in Sections 4--5 --- but most readers will want to start with
the journal article and use this document only for details and
leftovers.

\section{Related Research}

\begin{quote} {\it [This section includes some references not included in  the journal
      article, and provides more context for some others.]
} \end{quote}

\subsection{Phonetic and Phonological Correlates} \label{sec:correlates}

Various phonetic and phonological correlates of reduction have been
identified:
\begin{enumerate} \setlength{\itemsep}{0em}
\item Short duration, specifically word durations that are shorter
  than the average for the word
  \citep{jurafsky1998reduction,kahn2015}, or vowels shorter than
  average duration for that vowel \citep{turnbull2015}.  

\item Centralized vowels, ``less dispersed vowels'', compression of
  the vowel quadrilateral, or, more generally, changes in vowel quality
  \citep{jurafsky1998reduction,turnbull2015}.

\item The application of specific phonological rules in various
  languages \citep{aguilar1993,picart2014analysis,machac-fried23},
  such as coda-consonant deletion in English
  \citep{jurafsky1998reduction}.

\item Intonational reduction, including deaccenting, lower F$_0$
  peaks, and steeper spectral tilt \citep{consonant-reduction,burdin2015,turnbull17}.

\item Elision of segments \citep{koreman2006perceived}, which can be a
  phonological rule, or alternatively, sometimes can be seen as an
  extreme form of duration reduction.

\item Increased coarticulation, to the point of, using the terms of
  \citep{niebuhr2011perception}, reduction down to underlying ``articulatory
  prosodies.''

\item  Articulatory undershoot \citep{gahl2012}.

\item  Desynchronization of articulatory gestures \citep{machac-fried23}.
\end{enumerate}

The relation among these factors is not well understood.  On the one
hand, some are clearly related. For example, people speaking faster
necessarily reduce durations and, at extreme rates, many aspects of
precision. On the other hand, these factors are probably not reducible
to one or two fundamental causes, as we know that the various
indicators are not always correlated
\citep{schubotz15,zellers16b,cohen-priva2023}, that reduction patterns may differ
substantially across languages
\citep{ernestus2011introduction,malisz2018dimensions}, and that there
may be different types of reduction
\citep{ernestus2011introduction,turnbull17}.

Further ideas for possible correlates can be obtained, indirectly, by
considering the properties of the likely opposites of reduced speech,
such as locally prominent words or syllables, and speech that is
unusually intelligible or clearly articulated, including, at the
extreme, hyperarticulated speech, Lombard speech, and ``clear
speech.'' These have been variously found to have higher F$_0$,
stronger harmonicity, and shallower spectral tilt or, concomitantly,
more energy in the high frequencies
\citep{beechey18,bradlow2002clear,picart2014analysis,
  niebuhr-reduction, lu-cooke,
  ludusan21,wagner-prominence,gustafsoncasual}.

To investigate this needs an understanding of what measurable features
correlate best with perceptions of reduction.  No such studies  seem
to have been done for dialog data, or indeed, at all.  Accordingly,
our first research question is: Which acoustic features correlate
best with perceived reduction in dialog? (RQ-A)


\subsection{Methods of  Investigation}

Perceptions of reduction tend to be weak and variable, so
investigations accordingly require tools and systematic procedures.

Ideally there would be general-purpose methods for automatic
estimation of the degree of reduction from speech signals, but current
methods are limited.  In principle one can apply a speech recognizer
to a signal and see where it fails, or where its output has low
confidence \citep{tu2018investigating,lubold19}, but in practice this
seems to have been done only for data where transcripts exist.
``Articulation entropy'' as an acoustic measure and measurements of
articulator displacement using electromagnetic articulography both
have promise, but their precision for small samples or regions of
speech is unknown \citep{sungbok-lee06,jiao-articulation}.  More
recent work has applied deep learning to the detection of reduction in
learner's speech \citep{lei-chen22}, although this has so far been
evaluated only on clear cases.  While there are prosodic correlates of
phonetic reduction, these seem inadequate for building a reduction
detector \citep{reduction-techreport}.

Given these tool limitations, all detailed studies of reduction have
so far required substantial human effort.  To briefly survey the
common methods:

First, one can exploit the negative correlation between
intelligibility and reduction \citep{ernestus02}, by using subjects'
ability to recognize words in isolation, extracted and heard without
their original contexts, as an index of reduction
\citep{machac-fried23}.  This doesn't scale easily, and is thus mostly
suitable for hypothesis-driven research, for example when using
controlled speech or small-scale corpus data.

Second, one may use word duration, as computed from transcripts, as a
proxy for reduction.  Specifically one may consider reduction to be
present when a word occurrence has a duration that is shorter than the
average for the word \citep{jurafsky1998reduction,kahn2015}.  This
method, however, does not tell the entire story, as the duration is
not a reliable proxy for other measures of reduction or for perceived
reduction.

Third, one may hand-label corpora for various specific correlates of
reduction.  These include reduced phoneme durations, more centralized
vowels, more co-articulation, and the application of various
language-specific phonological rules, such as coda-consonant deletion
in English \citep{jurafsky1998reduction,turnbull2015,
  koreman2006perceived,aguilar1993,picart2014analysis,machac-fried23,
  niebuhr2011perception}.  One may also use corpora which have both
full segmental labels and canonical transcriptions, from which one can
infer where reduction occurs
\citep{jurafsky1998reduction,niebuhr-tt}. The problems here are that
hand labeling of course does not scale, and that no single correlate
may adequately proxy for all reduction phenomena \citep{burdin2015}.

Overall, there currently exist no generally usable methods for
estimating the phrase-level degree of reduction in speech, let along
dialog speech.  Thus our second research question: can we easily
develop a tool for the automatic detection of reduction? (RQ-B)

\subsection{Other Interesting Related Research} 

Reduction is uncommon in foreigner- and infant-directed speech; on the
contrary, these often involve hyperarticulation.  Interestingly, the
hyperarticulation in in foreigner-directed speech may be a reason why
it can be perceived, out of context, as conveying negative affect
\citep{uther2007}.  A lack of reduction may be characteristic of the
speech of some autistic adults, whose vowels tend to have greater
articulatory stability \citep{kissine2021}.

 One interesting investigation was that of \citep{picart2014analysis},
 in which the speaker was induced to speak less clearly by giving him
 feedback in the form of ``an amplified version of his own voice.''

It has been noted that common reductions are things that language
learners do not acquire without effort, and need to be taught
\citep{reduced-efl}. **

\section{Annotating for Perceived Reduction}

\begin{quote} 
{\it [Most aspects of the annotation are described in the journal article.
  This section adds various information, especially regarding the
  inter-annotator agreement. ]}
\end{quote}

Our aims for the annotation were: 1) Create data for evaluating
reduction detectors, whether as a classification or regression models,
2) Create data for training a reduction detector, 3) Perhaps, as a
side effect, gain qualitative insight into the functions and nature of
reduction, 4) Create data for measuring human inter-annotator
agreement, to use as a reference point for evaluating detectors.

\begin{table}[th]
\begin{center}
{\small \begin{tabular}{ccllcc}
    \multicolumn{1}{c}{Conversation} & \multicolumn{1}{c}{Annotated Range}  &&&
    \multicolumn{1}{c}{Conversation} & \multicolumn{1}{c}{Annotated Range}  \\
\hline
EN\_006 & [0:00 -- 11:04]\rule{0ex}{2.6ex}  &&& ES\_001 & [0:00 –- 6:00]\rule{0ex}{2.8ex} \\
EN\_007 & [0:00 –- 4:13] &&& ES\_003 & [0:00 –- 2:00] \\
EN\_013 & [0:00 –- 6:10] &&& ES\_008 & [0:00 –- 2:00] \\
EN\_033 & [0:00 –- 5:00] &&& ES\_012 & [0:00 –- 5:00] \\
EN\_043 & [0:00 –- 5:00] &&& ES\_022 & [0:00 –- 5:00] \\
          &                           &&&ES\_028 & [0:00 –- 5:00] 
\end{tabular}  }
\end{center}
\caption{Ranges over which reduction was labeled, for English (EN) and
  Spanish (ES) conversations. }
\label{tab:recording}
\end{table}

\begin{figure}[th]
      \includegraphics[scale=1.1, clip, trim =.2cm 1cm .2cm 1cm]{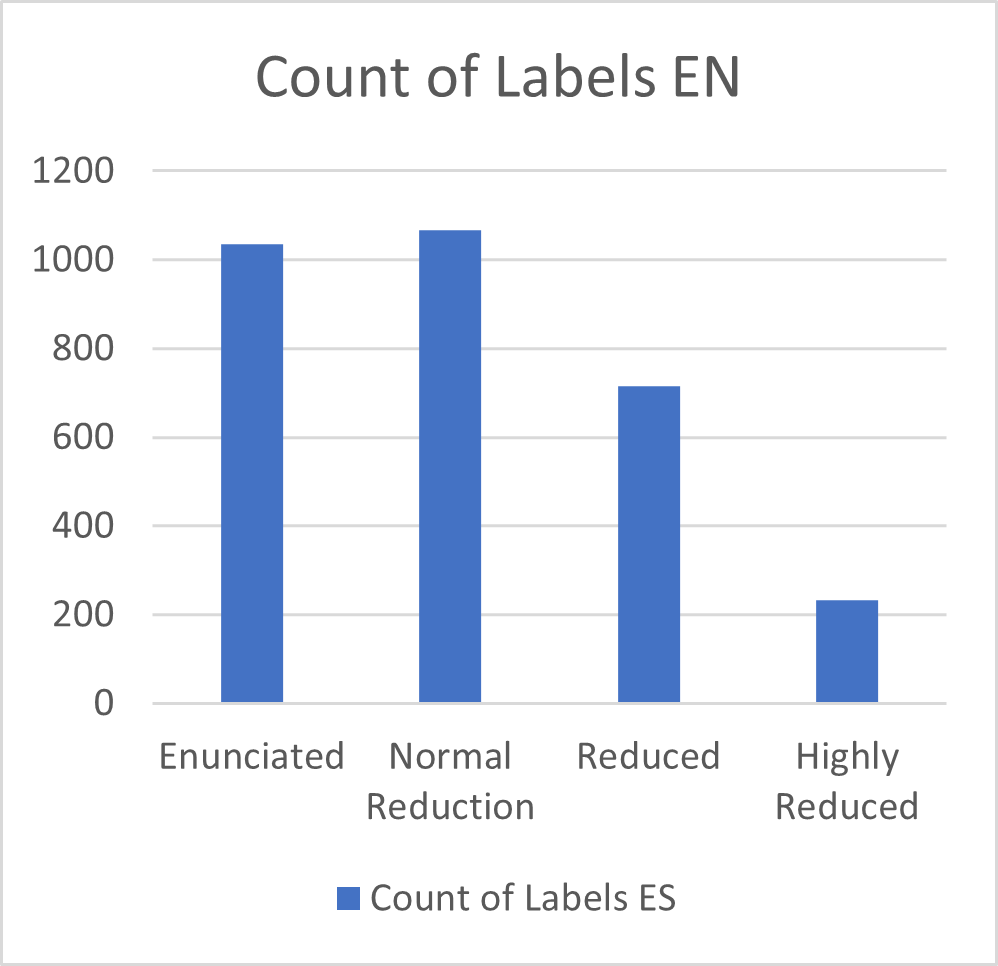}
      \includegraphics[scale=1.1, clip, trim =.2cm 1cm .2cm 1cm]{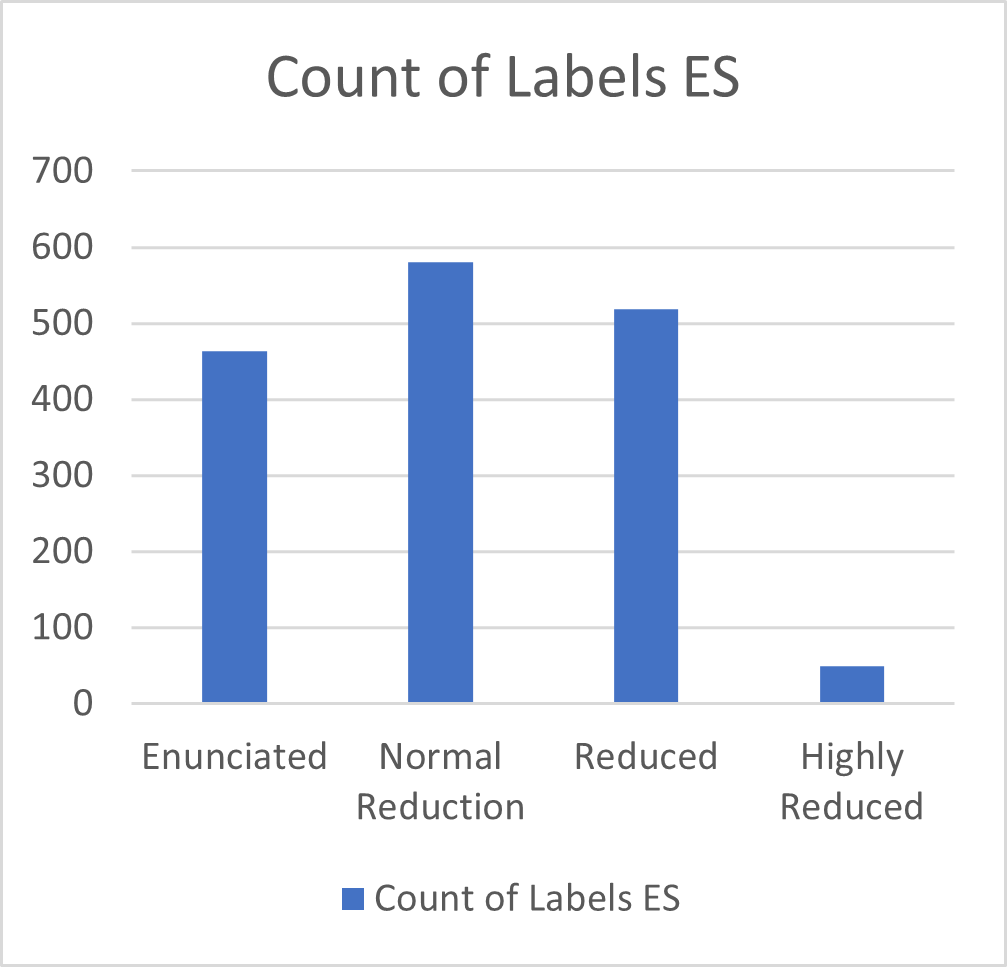}
      \caption{Count of labels for English (left) and Spanish (right)}
      \label{fig:counts}
\end{figure}

Table \ref{tab:recording} lists the conversation parts which were
annotated.
The first set of labels were unfortunately lost, so all were done a
second time, and these redone labels were the ones we used.
Figure \ref{fig:counts} show the counts of regions receiving each
label.

\subsection{Inter-Annotator Agreement}   \label{sec:iia}

The second annotator's training was brief: she received the guidelines
and went over some examples with the first annotator to see how he was
applying the labels.  To save time and simplify comparisons, she
labeled regions as already demarcated by the first annotator.

\begin{figure}[t!]
      \centering
      \includegraphics[scale=1.6, clip, trim = {5cm 1mm 6cm 5mm} ]{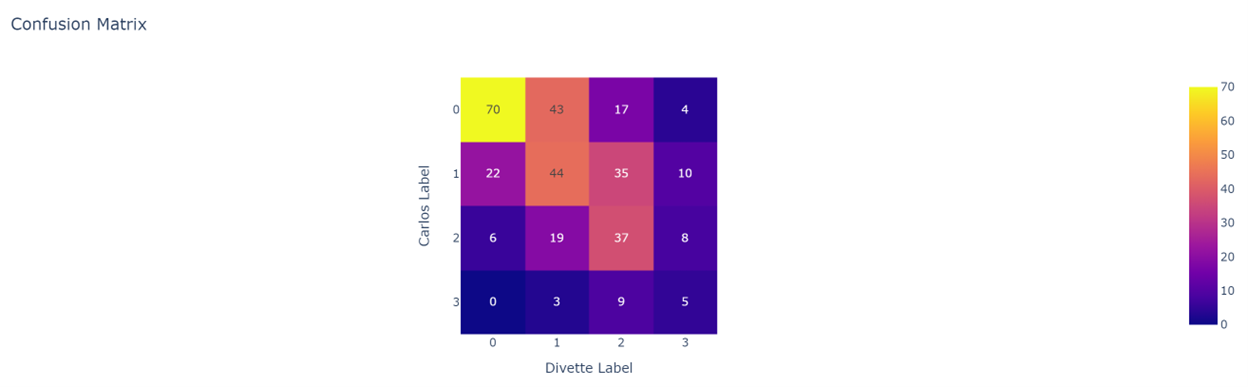}
      \caption{Inter-annotator confusion matrix, across both EN and ES labels}
      \label{fig:ia-confusions}
\end{figure}

From the confusion matrix, Figure \ref{fig:ia-confusions} we see that
agreement was moderate for the judgments of enunciation, weak for
judgments of normal articulation and mild reduction, and very poor for
judgments of strong reduction, which was also the rarest category.

To get a better idea of the reasons for the divergences, we examined
all cases in the English where the annotations differed by more than
1.  Numerous factors seemed  likely to have been involved, which we
can group into 6 categories:

\begin{enumerate} 

\item Varying Degrees of Reliance on Context.  Given an ambiguous
  sound, after you figure out what word it represents, it may be hard
  to set aside that information when relistening to the sound.  That
  is, after you know what the word is, it can seem like it was
  obviously there in the signal.  The following examples illustrate
  cases where this may have happened.  In each, the numbers indicate
  the annotations for the following word; for example, in the first,
  0/2 indicates that the annotations for {\it universes} were 0 and 2,
  by the first and second annotators, respectively.

\begin{itemize}
\item 
{\it I hear the argument a lot of times that Goku, ``oh, his powers
  only work because of, it's in-universe,'' but then again, you're
  just getting two people from /  from different 0/2 universes,
  though} (043@0:33)

\item  {\it for real, 3/1 for 3/1 real}  (043@1:02)
\end{itemize} 

In general, since the guidelines asked annotators to provide
"subjective judgments of being \ldots possibly hard to understand
without context," judgments could vary depending on the annotator's
ability to consider the hypothetical, setting aside the actual context
that they had heard.

\item Varying Effects of Prosodic Confounds and Correlates

Perception of reduction on specific words may be influenced by other
prosodic goings-on.   These may include:

A.  Emphasis, as in  

\begin{itemize}
\item   {\em \ldots she was \ldots ripping the skin, 3/1 like back} (006@6:29),
\end{itemize}

Here the phrase is the climax of a story, and that may have affected
the annotator's perceptions even of the word {\it like}.  (Here the 1
label is the post-hoc opinion of the first author, who noticed this
example in the course of examining different types of reiteration
during the ``other observations'' analysis described below.)

B. Lengthening,  as in 

\begin{itemize}
  \item {\it   Naruto after like twenty 0/2 four, episodes, I'm like}  (043@1:58)
\end{itemize}

C. Creaky Voice, as in 
\begin{itemize}
\item {\it    Okay, who do you think would, would Goku win, or would One Punch Man 0/2 win?} (043@0:12)
\end{itemize}
 where the last clause is very creaky and quiet.  (This speaker seemed
 to have a pattern of using creaky voice to downplay highly
 predictable information, as here, given that One-Punch Man had just
 been mentioned.)  This raises the question of how to consider creaky
 voice, which can impair intelligibility \citep{cammenga18}, but is
 not usually considered to be a form of reduction.

D. Falsetto, as in

\begin{itemize}
\item {\it 0/2 One 0/1 Piece 0/2 is real} (043@1:44) 
\end{itemize}

E. Laughing while Speaking, as in 

\begin{itemize}
\item {\it [laughter]  1/3 I,   I've   only   watched  like} (043@1:16) 
\end{itemize}

Possibly some of these differences may be side-effects of the
annotator's awareness of general tendencies regarding where reduction
appears: emphasized speech is often enunciated or at least not
reduced, and fast speech is often reduced while long-duration words
seldom are.  Since these speech aspects are probably more salient than
reduction, it would not be surprising if their perceptions affected or
even masked perceptions of reduction.

\item Varying Normalization Styles.  The guidelines did not specify
  whether the judgments were to be absolute or relative.  This was
  perhaps a reason why the second annotator's judgments tended towards
  the higher reduction levels.  The conversation that both annotated,
  DRAL EN\_043, featured two male speakers, neither of who were
  generally speaking very clearly, and in particular, the left speaker
  spoke quite fast and had a rather gravelly voice.  It is possible
  that the first annotator compensated for this, and used the
  ``reduced'' label, for example, to mean ``reduced more than typical
  for this speaker.''  (Previous work has shown that the perceptions
  of reduction can be ``modulated by the reduction level of the
  surrounding utterance context \citep{niebuhr2011perception}.)


When conveying pragmatic functions, reduction relative to the norm for
the specific speaker is probably more relevant than reduction in an
absolute sense, so it is fortunate for our purposes that the first
annotator appears to have  labeled relatively.

\item Variant Perceptions of Very Short Words. 
Judgments seemed to diverge more often for very short words:
{\em on}, {\em of}, and {\em be}, as in 

\begin{itemize}
\item  {\it  I hear the argument a  lot 1/3  of times, that} (043@0:24)
\end{itemize}

In two cases the annotator's difficulty was likely compounded by the
fact that the segmentation into regions was at too fine a level, and
for these words was slightly off.  For purposes of the analysis
below, these differences affected very few speech frames, so their
effect on the statistics will be negligible.

\item Local Comparisons. 
It seems possible that in one case the word {\it only} was judged to
be enunciated just because it was rather clearer than its neighbors,
rather than enunciated in a more absolute sense.  

\begin{itemize}
\item {\it  1/3 I,  1/2 I've  0/2 only 1/1  watched 0/1  like} (043@1:16) 
\end{itemize}

Local context effects on reduction perception are known to exist
\citep{niebuhr2011perception}.  This issue may also relate to that of 
differences in normalization styles, mentioned above. 

\item Idiosyncratic Labels. In a couple of cases, none of the above
  factors were obviously involved.  On the one hand, perhaps in these
  one annotator or the other simply made a data entry error.
  Certainly we know that noise can never be fully eradicated.  On the
  other hand, these could reflect real differences of perception, for
  as-yet unknown reasons.

\end{enumerate} 

In the course of reviewing these divergences, it became clear that
EN\_043 was an atypically challenging conversation --- no others
featured such extreme uses of creaky voice, falsetto, and laughter ---
so the level of agreement reported above likely understates what would
be found for more typical conversations.

We also noted that the differences of opinion affected at most one or
two words per phrase.  For the functional analysis below, which relies
only on average behavior, across multi-word regions and across many
such regions, such differences regarding specific words can be
considered to be just a minor noise source.


\subsection{Possible Revisions to the Guidelines and Procedure} 

Overall the guidelines seemed adequate for guiding the annotators to
generally do what we wanted, however, 1) The ``granularity'' clause
should probably be emphasized.  2) The ``confounds'' clause appears
naive, based on the results reported in the next section, and is
likely neither effective nor necessary.  3) Depending on the purpose,
it could be worth augmenting the ``criteria'' clause with the wording
"relative to that speaker's typical behavior."  4) As the label set
used, \{0, 1, 2, 3\}, could cause confusion, if the annotator has a
memory lapse and thinks of 0 as denoting 0 intelligibility, it may be
better to change the ``codes'' clause to specify a more memorable set
of labels, such as \{e, n, r, rr\}.

Regarding the procedure, one might make conservative changes or a
radical one.

The conservative changes might include: 1) A proper training sequence.
In particular, if correlate-influenced perceptions are a frequent
problem, this might include an explanation of the nature of emphasis,
creaky voice, falsetto, and so on, with examples, to help annotators
recognize these for what they are, rather than forms of reduction or
enunciation. 2) A restriction to one pass over the data.
Perfectionist annotators may wish to understand every word before
making any judgments.  The audio quality of these recordings being
mostly excellent, if one uses headphones and raises the volume when
necessary, almost every word can be heard, and, since these
conversations are quite interesting, the temptation to do so is real.
The problem is that it is impossible to unhear a word, so if reduction
judgments are made as a second step, they may be biased.  Accordingly
it may be useful to allow the annotator only one pass over the data,
and perhaps also  to forbid touching the volume knob.

The radical change would be to move away from subjective judgments and
instead just measure intelligibility, as the fraction of words
correctly recognized.  Decontextualized presentation would be
advantageous: for example, we could give annotators randomly selected
1-second spans.  Then reduction could be estimated as the fraction of
the words they report being unable to identify, or for which their
guesses do not match the reference transcript.

\section{Features and Correlations}

\begin{quote}{\it 
    [The topic of this section is not addressed in the journal article.] }
  \end{quote}

\subsection{Features Considered} 

To address questions RQ-A and RQ-B, we needed a set of acoustic
features. We limited attention to features that could be automatically
computed, and, for the sake of convenience, robustness, and
understandability, chose to select from a set of previously developed
features, designed to work robustly for dialog data and, in
particular, to include proper normalizations
\citep{midlevel-toolkit23}.

Most of these were selected based on  observations and
findings from previous research (Section \ref{sec:correlates})

\begin{description} \setlength{\itemsep}{0em}

\item[tl] a measure of how strongly the pitch is low in the
  speaker's range

\item[th] a measure of how strongly the pitch is high in the
  speaker's range

\item[vo] intensity

\item[cr] a measure of creakiness

\item[vf] voicing fraction

\item[re] a crude measure for vowel centralization, averaging over
  voiced frames the evidence for the cepstal coefficients being
  atypically close to the global average, which is presumably close to
  schwa.

\item[en] conversely, a measure of the extent to which voiced
  frames tend to be distinct from the global average cepstrum. 

\item[le] a very crude measure of lengthening, inversely
  proportional to the cepstral flux.  This tends to be higher when
  there is a lengthened vowel, for example.

\item[sr] a very crude proxy for speaking rate, measuring the
  average frame-by-frame differences in intensity.  This tends to be
  higher in times where vowels and consonants occur in quick
  succession.
  \end{description}

We also included features related to spectral tilt.  In this we were
inspired by the fact that spectral tilt, although a well-known
acoustic feature, and sometimes reported to be associated with
prominence, has not much been examined from a pragmatic functions
perspective, either directly or as a correlate of reduction.
Accordingly we developed several tilt features and added them to the
Prosodic Features Toolkit, to support investigation.  Specifically
these were:

\begin{description} \setlength{\itemsep}{0em}
\item[st] the average spectral tilt.   The implementation followed \citep{lu-cooke}: ``spectral
  tilt was computed via a linear regression of energies at each
  1/3-octave frequency."

\item[tr] the range of the spectral tilt within the window. 

\item[tf] a measure of the flatness of the spectral tilt, high when
  there is generally more energy in the high frequencies than typical.

\item[tm] a measure of when the spectral tilt is ``middling'',
  neither clearly flat nor strongly negative (steep).
  \end{description}

In addition we included four features without having any reason to
believe they would be relevant.

\begin{description} \setlength{\itemsep}{0em}
\item[np] a measure of the narrowness of the pitch range. 
\item[wp] a measure of the wideness of the pitch range. 
\item[pd] a measure of the disalignment between the pitch peak and energy peak, typically measuring late pitch peak on stressed syllables.
\item[sf] an estimate of the speaking fraction: the fraction of the time within the region devoted to speech versus silence. 
  \end{description}

These short descriptions are suggestive rather than accurate.  The
actual computations are designed to better match perception and to be
more robust. Fuller descriptions appear in the code and documentation
\citep{me-cup,midlevel-toolkit23}.

This feature set roughly covers correlates 1, 2, 4, and 7 noted in
previous research (Section \ref{sec:correlates}), but not correlates
3, 5, 6 and 8, at least not directly.

\begin{table}[tb]
  \begin{center}
    \begin{tabular}{|c|}
\hline
      --250 ms to --100 ms\rule{0ex}{2.6ex} \\
      --100 ms to --20 ms \\
      --20 ms to 20 ms \\
       20  ms to 100 ms \\
      100 ms to 250 ms \\
\hline
      \end{tabular}
  \end{center}
\caption{Spans for feature computations, where 0-10ms is the frame
  whose reduction level is being predicted.}
\label{tab:spans}
\end{table}

\begin{table}[t]
{\small
  \begin{center}
    \begin{tabular}{lcrr}
      base feature & span & English & Spanish \\
      \hline 
      pitch highness (th) & --250 $\sim$ --100 & & .087\rule{0ex}{2.6ex}\\
      ~~~~ '' & --100 $\sim$ --20  & & .103 \\
      ~~~~ '' & --20 $\sim$ 20     & & .098 \\
      ~~~~ '' & 20 $\sim$ 100      & & .106 \\
      ~~~~ '' & 100 $\sim$ 250 & .067 & .111 \\
      pitch lowness (tl) & --250 $\sim$ --100 & & --.068\rule{0ex}{2.8ex} \\
      pitch wideness (wp) & 100 $\sim$ 250 & & .071\rule{0ex}{2.8ex} \\
      intensity (vo) & 20  $\sim$ 100 &    & .078\rule{0ex}{2.8ex} \\
      ~~~~ ''  & 100 $\sim$ 250 & .083 & .109 \\
      low cepstral flux (le) & 20  $\sim$ 100 & .067 & .061\rule{0ex}{2.8ex} \\
      ~~~~ ''  & 100 $\sim$ 250 & .080 & .072 \\
      speaking fraction (sf) & 100 $\sim$ 250 & .077 & .093\rule{0ex}{2.8ex} \\
      tilt range (tr) & 100 $\sim$ 250 & .081 &  \rule{0ex}{2.8ex}
    \end{tabular} 
  \end{center}
  }
  \caption{Features with strong correlation with reduction.  Spans in milliseconds. }
  \label{tab:big}
\end{table}

Because we thought that some indications of reduction at given
timepoint might appear before or after that timepoint, we computed
each feature over windows spanning a half second around it, seen in
Table \ref{tab:spans}.

\subsection{Correlations} 

We used this feature set to address RQ-A.  Specifically, we computed
correlations with judgments of reduction for 10 millisecond frame in
every speech segment.

 Table \ref{tab:big} shows the features whose correlations
for English or Spanish had magnitude greater than 0.06, and Figures
\ref{fig:corr-en} and \ref{fig:corr-es}
show all the correlations.  We observe:
\begin{enumerate} \setlength{\itemsep}{0em}
\item The most highly correlating features related to pitch highness,
  and other relatively highly correlating features were tl (in
  anticorrelation), vo, wp, and le.

 All of these were contrary to expectation.  Fearing that these
 surprising results may have been due to a flaw in our methods, we
 re-listened to some of the dialogs, and confirmed that reduction was
 not infrequent at times where the speech was variously loud, slow or
 with pitch that was high or wide.

\item 
For many features, there was a tendency for  later spans to be more
informative.  For example, the pitch height over the span 100 $\sim$
250 milliseconds {\it after} the frame of interest was the most
informative feature. This may reflect the tendency in many utterances
for reduction to increase over time.

\item Overall the features correlated similarly in English and
  Spanish.

  \item However the tilt features correlated differently in the two
    languages: in English, middling spectral tilt correlated with
    reduction, but in Spanish, the range of spectral tilt was a good
    correlate.

\item There were no individual features that correlated strongly, with
  the highest being around 0.08 for English and 0.11 for Spanish.

\item 
The speaking fraction feature, sf, correlated positively,
possibly  because clear unvoiced stops are less frequent in reduced
speech or because inter-word pauses are rare.

\item 
The 're' feature, designed to capture vowel centralization, does
correlate positively with reduction, but only very weakly.  Unlike the
other features, for 're' the most informative span is the one centered
around the frame of interest.
\end{enumerate}

The most interesting finding is the first: the correlates of reduction
in dialog differ from those in read speech.

One possible explanation is that, in dialog, people not wishing to
show disinterest or disrespect, may compensate for reduced effort in
one respect, such as articulatory precision, with increased effort in
others, such as pitch height, pitch range, and intensity.  In these
conversations the speakers were always highly engaged, but of course
that is not true in general.

\section{Predictive Models}

\begin{quote}{\it 
[The topic of this section is not addressed in the journal article.] 
}\end{quote}

To address question RQ-B, regarding the prospects of developing a tool
for the automatic detection of reduction, we trained a few models.
For each, the task was, given a timepoint within an annotated region
and a half-second of context, to predict the annotator's reduction
label.

We wanted a tool that would not only correlate well with human
perceptions, but in addition be small enough to be easily deployable,
and be simple and explainable.  For this initial foray, we chose to
prioritize the last goal.  Thus, rather than trying solutions using
speech recognition or pretrained models, we only tried models built on
the features listed above.  




Each model was trained on all the annotated conversations for each
language except the last, namely EN\_043 and ES\_028, which were
reserved for evaluation.  Thus the training/test splits were
approximately 84/16 for English and 80/20 for Spanish.

Our performance metric was the correlation: a model was better to the
extent that its predictions correlate better with the annotated
values.

\begin{table}[th!]
\centering
\begin{tabular}{lrr}
      & \multicolumn{2}{c}{Correlations} \\
Model & English & Spanish \\
\hline
Linear regression & 0.243 & 0.168\rule{0ex}{2.3ex} \\
kNN & 0.014 & 0.012 \\
CNN & 0.061 &  \\
second annotator &.570 & \rule{0ex}{2.3ex} 
\end{tabular}
\caption{Correlations (Pearson) between predicted and annotated values}
\label{tab:results}
\end{table}

We built three models: linear regression model for simplicity, a k
nearest neighbors model expecting it to capture specific patterns that
linear regression would miss, and a convolutional neural network
expecting it to exploit more temporal context.  Code for all of these
is released at {\tt https://github.com/Caortega4/reduction-detection}
.

The results are  seen in Table \ref{tab:results}. We found  
\begin{enumerate} \setlength{\itemsep}{0em}
\item Surprisingly the linear regression models performed best, for
  both languages.
\item Surprisingly, reduction in English was more predictable than for
  Spanish, despite the higher feature correlations for the latter,
  although this might be an artifact of data size differences or 
  speaker idiosyncrasies.
\item Unsurprisingly, reduction is not realized the same way in the
  two languages: the prediction quality for models trained on one
  language were lower for the other language.
\item Performance was far lower than the topline (human) performance,
  even though the latter is probably an underestimate, since it was
  computed over labels, not frames, and only over one atypical
  conversation.
\item Overall the prediction quality was too low to be probably useful
  for most purposes.  
\end{enumerate} 
 
Thus our answer to RQ-B is that prediction of reduction directly from
such features is not easy.  Our idea was that, given various
literature reports of prosodic features that correlate with reduction,
these features together would be adequately predictive, however this
was not the case.  Thus we were unable to create a working reduction
detector.

We think this suggests that these reported correlations may not be
reliable or general.  Rather, they may be components of patterns that
also involve reduction, as for example various prosodic constructions
\citep{me-cup} that involve reduction and other features in various
temporal configurations.  The varying correlations reported between
reduction and various prosodic features may simply reflect the varying
prevalence of specific prosodic constructions in the various genres
studied, rather than being direct correlates of reduction instead. 

If this is correct, only minor improvements in prediction quality can
be expected from the obvious tweaks to this modeling approach:
training on more data, using different features, using more features,
better handling of context-feature computations (in cases where a
contextual feature overlaps silence), and using more sophisticated
modeling.  Rather, the best way to improve on these results will
likely involve using speech recognition results or using pretrained
models.

\section{More on the Functional Annotation Process}

\begin{quote}{\it 
[The functional analysis had four phases.  To keep the story simple
  the journal article highlights only the fourth, namely the
  independent annotation phase; here we discuss the others.] }
  \end{quote} 

As a preliminary note, our original intention was to use a reduction
detector to compute the degree of reduction everywhere in speech, and
then feed in this as an additional feature for our standard
prosody-analysis workflow \citep{ward-breathy}.  Doing so would have
enabled us to discover how reduction occurs in patterns with other
prosodic features, and then to study the functions of such patterns.
Perhaps we could have used the hand-labeled reduction labels in such a
workflow, however, lacking much data, we instead opted for simpler
methods.

\subsection{The Initial List}   \label{sec:initial-list}

We started by developing a list of pragmatic functions that commonly
occur with reduction.  To do this, the second author examined all
regions that were labeled reduced in ten minutes of dialog (minutes
9--11 of DRAL EN\_006, 0-2 of EN\_007, 0--5 of EN\_013 and 3--4 of
EN\_033).  He noted down the functions observed and grouped them into
categories, refining the categories as he examined more instances,
eventually arriving at nine categories.: 1) fillers, interjections,
and backchannels, 2) prosody carriers like {\em like} and {\em you
  know}, 3) uncertainty markers, 4) recapitulations, 5) predictable
words, 6) downplayed phrases, including parentheticals, 7) topic
closing moves, 8) turn grabs, 9) personal feelings.  These are also
listed briefly in Table \ref{tab:first-pass-english} and discussed in
full in Section \ref{sec:more-english}.

\subsection{The First-Pass Annotation}

The same author then annotated all the non-examined dialogs for the
presence of these 9 functions, plus two controls.

Specifically, to determine where these functions were present, the
first author annotated their presence in 21 minutes of English dialog,
namely all the data not used in developing the initial list.  Again,
both left and right tracks were annotated.  This annotation was not
blind to the hypotheses, but was at least done without knowledge of
the second author's reduction annotations.  In addition, as controls,
two additional functions were labeled, neither expected to correlate
with reduction, namely positive assessment and negative assessment.

Since reduction is common, any pragmatic function will likely
sometimes overlap reduced regions, just by chance.  We imagined,
however, that the connections noted above would be significantly
associated: that each of these nine functions will overlap reduced
regions more often than chance.  We note that, while the reduction
annotations are not entirely reliable, none of the complicating
factors (Section \ref{sec:iia}) seem likely to much relate to any of
the pragmatic functions under investigation, so for this analysis we
took the annotations as the truth.

We tested the strength of association for each function by testing
whether regions marked with that function have a higher mean reduction
level than the global mean over all speech frames, 0.98.  For this we
used a one-sided t-test.  Table \ref{tab:first-pass-english} shows the
results.  The overall idea of of a connection between reduction and
pragmatic functions was supported, as were connections for four of the
specific functions.  Some of the effect sizes were fairly large, given
that the overall standard deviation was 0.89 steps.  Thus, for
example, the effect size of being part of an uncertainty marker was
0.66 standard deviations.  This was more a hunting expedition than a
test of specific hypotheses, since we had no strong reason to believe
that any of these functions was reliably associated with reduction,
but in the table we do mark the functions with $p < 0.05$.  With nine
functions identified, there were 9 independent tests.  However,
because these nine tests are a ``family'' in the sense of providing
support for the overall claim that reduction conveys pragmatic
functions, to evaluate the strength of the latter claim we also tried
a Bonferroni correction: one function survived.



\begin{table*}
\begin{center}
\begin{tabular}{lrrrl|rrrr}
&&&&   &\multicolumn{4}{c}{percent with reduction label:}  \\
  & average &\multicolumn{1}{c}{n} & \multicolumn{1}{c}{$p$}  & &  \multicolumn{1}{c}{0}
  & \multicolumn{1}{c}{1} & \multicolumn{1}{c}{2} & \multicolumn{1}{c}{3} \\
\hline
all regions & 0.98 & 3051 & \multicolumn{1}{c}{--} & &35\%&38\% &21\% &6\%  \rule{0ex}{2.6ex} \\
\hline
Uncertainty Markers\rule{0ex}{2.6ex}& 1.57 & 23 & $<$.001&*+ & 16\% & 27\% & 41\% & 16\%\\
Topic Closings 	    & 1.81 & 7  & .011&* & 0\% & 42\% & 35\% & 23\% \\
Turn Grabs          & 1.72 & 8  & .026&* &  8\% & 40\% & 23\% & 29\% \\
Predictable Words\rule{0ex}{2.6ex}   & 1.49 & 11 & .036&*  & 13\% & 38\% & 36\% &  13\%\\
Personal Feelings   & 1.13 & 38 & .068& \textdagger & 25\% & 42\% & 26\% &  6\% \\
Downplayed Phrases  & 1.32 & 11 & .093&  & 24\% & 31\% & 35\% & 10\% \\
Prosody Carriers\rule{0ex}{2.6ex}   & 1.16 & 40 & .121 & & 34\% & 29\% & 24\% & 13\%\\
Recapitulations     & 1.04 & 51 & .266 & & 36\% & 36\% & 17\% & 12\% \\
Fillers, etc.       &  .75 & 25 & .896  & &55\% & 14\% & 30\% &  0\% \\
positive assessments  \rule{0ex}{2.6ex}        &  1.34 & 27  & .005& \multicolumn{1}{r|}{* !}&   24\% & 29\% &37\% & 10\%\\
negative assessments  &  0.86 & 43 & .884 & \multicolumn{1}{r|}{!}& 48\% &25\% & 20\% &7\% 
\end{tabular}
\end{center}
\caption{Reduction Statistics for Various Pragmatic Functions in
  English, First Pass. The second column indicates the average
  reduction levels, with * indicating those whose t-tests came out
  with $p < 0.05$, + indicating the one that looks significant even
  after Bonferroni correction, \textdagger indicating one nearly
  significant, as discussed in Section 6.4, and !  indicating {\it post
    hoc} tests done later for functions that were originally intended
  as controls.  The third column shows the number of occurrences,
  counting function-labeled regions that overlapped at least one
  reduction-labeled region.  Functions are ordered by strength of
  relation to reduction, as measured by $p$ values, in the fourth
  column. The remaining columns show the percentages of 10-millisecond
  frames at each reduction level.  The ``all regions'' statistics are
  for all regions with reduction labels, not limited to those at times
  for which functional labels were assigned. }
\label{tab:first-pass-english}
\end{table*}

However there were several flaws with this work.

Two related to the annotation process. First the annotator was aware of
the purpose of the annotation, namely to study the functions of
reduction. Second, in order to simplify the annotation process,
regions that related to more than one of the nine categories were
annotated with only the first that applied, using the order in Section
\ref{sec:initial-list} This was quick and convenient because the list
is ordered so that the earlier ones are generally easier to judge than
the later ones, but led to slight underestimation of the relations
between reduction and the functions lower on the list, as discussed
below. Accordingly we did a second-pass of annotations, as described
in the journal article.  There the annotator was naive to the aims of
the study, and avoided the overlap problem by creating multiple Elan
tiers per track, as many as needed to enable annotation of all
simultaneously present pragmatic functions.

The other flaw was intrinsic to the use of corpus data: thee results
could be affected by many uncontrolled sources of variation, such as,
perhaps an increased frequency of function words in positive
expressions, or the possible tendency for people to emphasize the
positive, and so repeat themselves more when doing so, and accordingly
speak faster.  For this reason we did the controlled experiment.

\subsection{Additional Listening}

The first author then listened to a number of the examples for each
type, trying to better understand why each category did or didn't have
a strong statistical correlation.

\subsection{The Second-Pass Annotation}

This was done as described in the Other Pragmatic Functions section of
the journal article.

\subsection{Comparison} 

For many of the functions, the results were different between the the
first-pass and second-pass functional annotations.  This was to be
expected, as the definitions are vague and the perceptions subjective.
Overall, the second annotator seemed to apply the criteria for each
category more loosely.  This may in part have been because she was
more sensitive and thus able to pick up subtle meanings that the first
annotator had missed, and in part because she was unfamiliar with the
usual terminology for describing conversation functions.  Clearly, in
future, training for functional annotation should be done better.

Such global considerations aside, the two annotators together reviewed
annotations of the two functions for which the results were most
different, namely Uncertainty Markers and Turn Grabs, both of which
had a strong association in the first pass but weak ones in the
second.  Specifically, they listened to all regions that she had
labeled with either of these tags in EN\_006.

\section{More on the Pragmatic Functions of Reduction in English} \label{sec:more-english}

\begin{quote} 
{\it [The journal article focuses only on the functions for which
    there was good support, namely positive assessment and topic
    closings.  This section gives more detail on functions for which
    the evidence was weaker, or didn't support the hypothesis at all.
    These comments are based on all the observations and speculations
 from across all phases of the analysis.]}
\end{quote}

In the illustrations, underlining indicates the approximate extent of
the reduced regions, slashes indicate speaker changes, and asterisks
mark the speech region that illustrates the point being made.  Audio
for all examples is available at {\tt
  http://www.cs.utep.edu/nigel/reduction}

\subsection{PO: Positive Assessment}

This category was included in the list of functions to annotate, not
due to any expectation of a connection with reduction, but with the
idea that it would serve as a control.  However in fact it turned out
to be strongly associated with reduction.  Examples include:

\begin{itemize}
\item {\it I want to work with the inmate population / *\underline{oh wow, that's interesting}} (006@1:53) 

\item {\it Yeah, because I got the research position here, and I \underline{thought, it was a} good, \underline{opportunity, because I want to} do my Ph.D.} (006@2:39)

\item{\it I was in taekwondo \ldots no, but. \underline{It was, it was pretty} cool.  I liked it.} (033@0:15)  

\item  {\it I've seen Bleach, I've seen Akame ga Kill / \underline{That one's} so good; I love \underline{that one.  That one's} like, a \underline{nice} short one ... }  (043@2:28)       
\end{itemize} 


\subsection{FI: Fillers, Interjections, and Backchannels}

These are mostly non-lexical items, for which the standard concept of
reduction does not apply.  However the occasional lexical items in
these roles, such as {\it well}, {\it really}, and {\it and}, were
often reduced.

\begin{itemize}
\item {\it \ldots other classes.  *\underline{And, like}, all my friends \ldots}  (013@1:52)
\item \ldots {\it it's like cloud services, it's like AWS / *\underline{oh, okay} / \underline{that kind of stuff}} (013@0:55)
\end{itemize} 

Statistically, however, this category showed up to be generally not
reduced.  Further listening revealed that this was in part due to the
frequency of elongated fillers such as {\em so} or {\em the}, where
the lexical identity was very clear.  However for these items
articulations did not seem particularly precise, suggesting again that
``reduced-enunciated'' is likely not a single dimension of variation.

\subsection{PC: Prosody Carriers}

This was a term we invented to refer to the words {\it like} and {\it
  you know} in cases where they convey no specific semantics.  Such
uses often seem to serve mostly to provide phonetic material to fill
out some prosodic construction.  Examples include
\begin{itemize}
  \item {\it all the summer classes I've taken, I've taken the whole summer.  That's fortunate for me, *\underline{you know}} (013@1:10)
  \item {\it and then you'll do some stuff with *\underline{like} numpy} (013@2:44)
\end{itemize}

 Prosody Carriers in general tend to be reduced, but when the word
 {\it like} appears as a filler, it is often somewhat lengthened and
 not reduced.

\subsection{UC: Uncertainty Markers}

These include words and phrases such
  as {\it I don't know} and {\it hopefully}.

\begin{itemize}
\item {\it maybe I'll be a TA; *\underline{I'm not really sure yet}} (013@0:36)
\item {\it yeah *\underline{I feel like I} still need more time to even figure out ...} (013@3:59)
\end{itemize}

While Uncertainty Markers were not invariably reduced, the tendency
was very strong.  These included such phrases as {\it I want to say},
{\it I think}, {\it you could say}, and {\it I guess}.

In the comparison phase, we found that the second annotator had often
taken {\it like} to be indicating uncertainty, where the first
annotator considered most such cases to be just discourse markers.
Further, the second annotator marked as uncertainty some cases where
the speaker was producing false starts and hesitations, where the
first annotator generally didn't.

\subsection{RE: Recapitulations}  In these the speaker reiterates a previous
  point, often using different words.

\begin{itemize}
\item {\it Well, I'm planning to go to New York, I'm taking an internship with Amazon / that's cool! / yes, *\underline{I'm going there} } (013@0:17)
\end{itemize}

Recapitulations, based on the literature, would be expected to
frequently reduce, but statistically the tendency was weak.  However
further listening revealed that recapitulations are not a unitary
category, as was already known for repetitions and repairs
\citep{zellers-why-rethink}.  Simple repeats of previous words, for
example when recovering from a false start or an interruption, may
indeed be generally reduced.  However words reused in a subsequent
phrase to reemphasize a point are often not reduced, nor, perhaps,
 are repeats in which one speaker echos a few words of the other, as a
 form of backchanneling.

\subsection{PW: Predictable Words}

In these the predictability may come
  from specific knowledge,
  or more shallowly, from collocation tendencies or the local
  syntactic context.
\begin{itemize}
\item {\it what are your plans for after the semester's *\underline{over}?}  (013@0:08)
\item {\it (listing his Fall classes) I'm taking Computer Security, *\underline{I'm taking uh} \ldots Machine Learning ...} (013@1:18)
\end{itemize}

\subsection{DP: Downplayed Phrases}

Speakers sometimes include
  parentheticals or side comments that are not intended to be
  responded to.  Some repair markers may fall in this category. 
  
  \begin{itemize}
  \item {\it [I'll be taking] Deep Learning, and, Advanced Algorithms (*\underline{which, that's supposed to be really hard}). How about you?} (013@1:26)
  \item {\it \underline{uhm}, I'm *\underline{I'm going to visit some friends right after finals, but then I'm}  taking a class. } (013@0:28)
    \item {\it I'm taking Computer Security, {\it I'm taking} Machine Learning *\underline{no, sorry, uh} Deep Learning \ldots.} (013@1:24)
  \end{itemize}

Downplayed Phrases turned out to be a diverse category,
  including repair markers, asides, self-talk, and parenthetical
  comments.  In general their degree of reduction {\it versus}
  enunciation may reflect the degree to which the speaker expects the
  hearer to ignore this phrase {\it versus} pay attention to it.



  \subsection{TC: Topic Closing Moves}
  
Closing out a topic may involve not
  only reiterations and downplayed phrases, but also cliched phrases
  and other ways to show that the speaker has nothing more to say
  about a topic.

  \begin{itemize}
\item \ldots {\it it's like cloud services, it's like AWS /   \underline{oh, okay} / \underline{that kind of stuff} yeah   \underline{yeah, you know, I don't know, this is what happens I guess.}}
    (013@0:57)
  \end{itemize} 
  
Strikingly, as seen in the table, Topic Closings were not only usually
reduced, they were never enunciated.

  \subsection{TG:   Turn Grabs}

  Sometimes a speaker takes the turn before
  knowing quite what he wants to say.  Also in this category we
  include ``rush-throughs'', where a speaker seems to revoke a turn
  yield by speaking quickly to forestall the other from taking the
  turn after all.

  \begin{itemize}
\item {\it nice.  \underline{So you're just going to be spending}, staying the whole summer there?} (013@0:20)
\item {\it we just barely know how to actually code. \underline{Because you're taking} Objects too now right?} (013@3:46)
  \end{itemize}

Incidentally, while Turn Grabs were often false starts and thus
``reparanda'' in the sense of \citep{shriberg-ipa}, in this corpus
these were generally not errorful statements that needed to be
corrected, but rather things that were said quickly to grab or hold
the floor, and turned out not to be formulated in such a way as to be
easily continuable.

Regarding Turn Grabs, the second annotator found vastly more
instances.  The main reason was that she used this tag for many
instances of supportive simultaneous talk.  She also used this for
various cases where the speaker suddenly sped up and produced a new
talk-spurt.

\subsection{PF: Personal Feelings}

These include preferences and desires.

  \begin{itemize}
  \item {\it Are we going to go eat pasta?  / I *\underline{don't really want} pasta right now} (007@1:05)
  \item  {\it   \underline{I mean, hopefully, I don't know, but. I don't know
    but. Like, *I don't want to } be inspirational, I just want to do
    something.  You know, \underline{like *I don't want to \ldots}}  (006@9:54)
  \end{itemize}  

Personal Feelings only narrowly missed a statistically significant
association with reduction, and only due to our poorly conceived way
to simplify the annotation process.  Specifically, as PF was the last
item on our list, its presence was only annotated when no other
functions had been identified over that span.  As a retroactive
attempt to compensate for this bad decision, we scanned all regions
labeled TC and PF, and found that two of the former and one of the
latter otherwise met the criteria for a PF label.  Had these three
been included in the annotations, personal feelings would have been
identified as associated with reduction at $p<0.033$.  Incidentally,
on further listening, the personal feeling category turned out to
diversely include not only immediate and current statements of simple
wants and preferences, but also thoughtful statements, variously
introspective, retrospective, or analytical.  This diversity may be
the reason why the second annotation was different enough that that
personal feelings, by those labels, tend not to be reduced.  Clearly
this category needs to be more carefully defined, and perhaps split
into subcategories.




%

\subsection{Speculative Observations}

In this data the reduction is generally on the subtle side: there are
no extreme mumbles as one might expect in cases of disengagement or
sleepiness, for example.  Almost none of the reductions in this data
were strong enough to impinge on intelligibility, even for the second
author, a non-native speaker. This suggests that different degrees of
reduction may have qualitatively different functions, that is, that
weak reduction may function quite differently from strong reduction.

Although the first phase found evidence that at least 5 functions
involve reduction, there is no reason to think they are generally
confusable.  Rather, it seems likely that reduction by itself is not
by determining the meanings, but doing so in concert with factors of
the local context, and, very likely, other co-occurring prosodic
features.  These may include: for uncertainty markers, slow speaking
rate; for topic closings, slow rate, long pauses, and low volume; for
turn grabs, fast rate and high pitch; for predictable words, low pitch
and low volume, and for positive assessment, fast rate and loudness.

\subsection{Functions of Nonreduction} 

While our main interest is the functions of reduction, this subsection
briefly treats the functions of enunciation and the absence of
reduction.

Just after initial analysis phase, the first author took another
listen to the first two minutes of DRAL EN\_013, this time paying
attention to regions where reduction was saliently absent: either
where the speech was clearly articulated or noticeably least not
reduced.  Again using qualitative induction methods, he identified the
following categories:
\begin{enumerate} \setlength{\itemsep}{0em}
\item new information
\item topic elaborations and transitions, such as with {\it how about you?}
\item completing an interlocutor's sentence
\item direct questions that invited long answers,  as in {\it what classes are you taking over the summer?}
\item confirmation questions
\item floor holds 
\item reported speech
\item direct answers 
\item facts (versus opinions)
\end{enumerate} 

Further, as seen in Table \ref{tab:first-pass-english}, two categories
turned out to tend to be non-reduced. 

Negative Assessment was common in these dialogs, and very often
exhibited enunciation.  With enunciated regions marked with $<$ and
$>$, examples include
  \begin{itemize}
\item {\it like super dan--, well, $<$dangerous$>$, quote-unquote, well, it was dangerous}
(006@7:47)

\item {\it I also learned a lot about what $<$inmates, or mentally ill individuals$>$ are $<$willing to do to their$>$ own $<$bodies$>$ } (006@4:22)

\item {\it yeah, I reached my,  $<$I peaked$>$ back then, just $<$all downhill$>$ from} (033@0:11) 
\end{itemize}

\section{More on the Experiment}

\begin{quote}{\it [The section gives detail, and describes the one aspect of the experiment that was omitted from the paper.] 
}\end{quote}

\begin{figure*}[thb]
\centerline{
\includegraphics[scale=0.58, trim =0cm 0cm 8.7cm 4.4cm, clip=true]{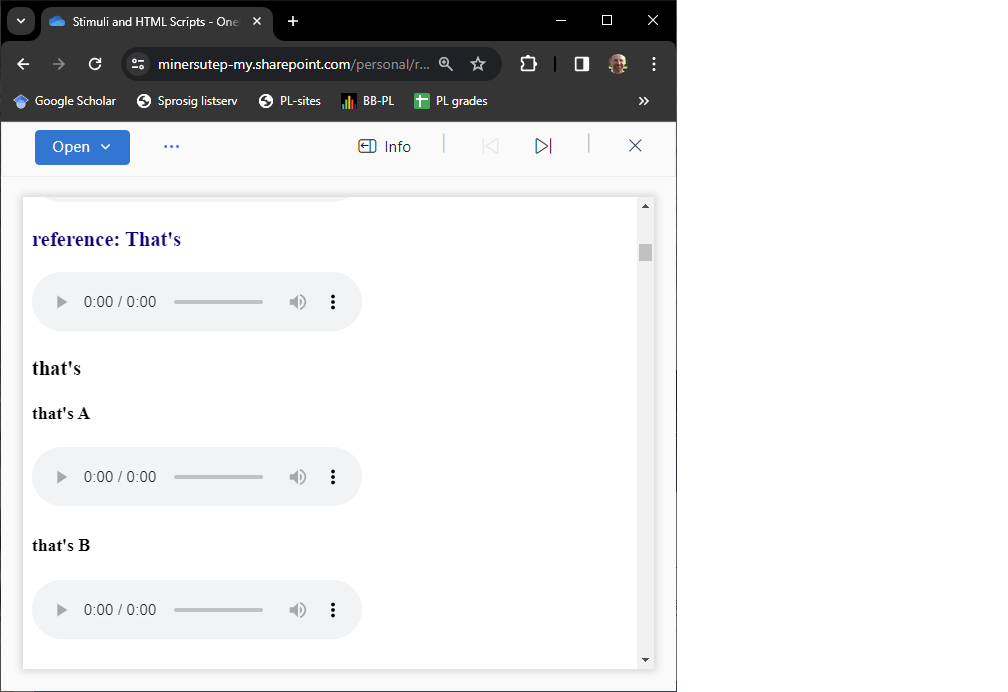}}
\caption{Stimulus screen. For each word, subjects could play the
  reference and the first and second conversational renditions.}
\label{fig:ssScreen}
\end{figure*}

Judges accessed the stimuli through screens like that in Figure
\ref{fig:ssScreen}.

In a brief investigation of reenactor effects, On one extreme we
found that Reenactor \#12's productions were strongly aligned with the
hypothesis; this was the one who spent the most time on his
productions.  On the other extreme, Reenactor \#1's productions were
judged opposite to the predicted direction.  It is not surprising that
the association between reduction and positive assessment, while
prevalent, is not valid for all speakers of English.  Possible factors
relevant to Reenactor \#1 include the fact that her positive
utterances seemed to differ mostly in pitch from the neutral ones, the
fact that she is a person with a generally precise speaking style, or
the fact that she was the first person we recorded, so our nervousness
may have prevented her from relaxing and producing truly
conversational-style utterances.  In future, we will probably follow
previous work \citep{niebuhr-michaud} in having the re-enactors
produce the utterances as part of skits performed with confederates.

In a brief investigation of judge effects, while the judgments of all
6 were in the predicted direction, the strengths of the effect varied,
from \#5, whose judgments were near chance, to \#22, the one who had
devoted the most time to the task, who overwhelmingly judged the
neutral phonemes to be clearer (101 to 49).

Early in the process, we formed a secondary hypothesis: that, within positive utterances,
the non-prominent words would be more reduced than the prominent ones.
This hypothesis arose from the observation that valenced adjectives
and verbs, such as {\it good, cool, I love, I liked it}, were often
non-reduced even when other words in the positive assessment were
reduced, and often these words seemed also to be prosodically prominent.

Not having a good idea of how to define prominence, nor any reason to
think that the actors would all place prominence in the same places,
we let the actors judge this themselves.  Thus, after all their
recordings were complete, each actor listened to each of their
positive and neutral recordings, and identified in each the word
``that sounds most prominent or stressed, if any." If an actor
identified any payload word as prominent in either the neutral or
positive rendition, we tagged it as such.  On average 28\% of the
words were tagged this way.  (For this, we counted AWS as 3 words.)
The actors were not consistent in their judgments, but all judged the
word {\it summer} as prominent, and other words frequently tagged
prominent were {\it interesting}, {\it actually} and {\it AWS}.

We found that people often do not agree on what is prominent, at leats
between us and our actors.

For the phrase where the duration difference was most confounding,
Actor \#12 marked {\it summer} as prominent in this phrase, and for
that word, there was 1 neutral-clearer judgment and 7 positive-clearer
judgments.

\begin{table}
  \begin{center}
    \begin{tabular}{lrr}
      &prominent & non-prominent \\
      \hline
      positive & 124  & 263 \\
      neutral & 183  & 330 \\
      total & 307  & 593
    \end{tabular}
  \end{center}
  \caption{Phonemes judged clearer.  (Of the positive-negative phoneme
    pairs, the numbers that were judged clearer.) }
  \label{prominence-effects}
  \end{table}

Regarding the secondary hypothesis, Table \ref{prominence-effects}
separates out the judgments on the prominent and non-prominent
words. Contrary to the hypothesis, the phonemes in prominent words
were actually slightly more likely to be reduced when positive than in
non-prominent words, although the difference is far from significant
(chi-square test, p $>$ 0.25).

Seeking to understand this better, we listened to a sampling of the
actors' productions.  Rarely was any word clearly prominent, and our
ideas of which words were prominent seldom agreed with those of the
actors themselves.

From this, we conclude that this hypothesis was ill-formed, and that
in future any studies of prominence in dialog should start with a
pilot study to see whether the concept can be operationalized.

Nevertheless, in the post-hoc analysis of the per-phoneme tendencies,
there was something going on with some of the words that caused
judgments contrary to the general tendency.   Perhaps these could
still be explained with some notion of ``prominence,'' if properly
defined. Specifically, there are pragmatic and prosodic properties of
three words which might be loosely described with this term.  1) The
term {\it AWS} was entirely new information when said in the corpus,
entirely unpredictable from the context.  Prosodically, the
reenactor's productions, following that of the corpus generally seemed
to be high in pitch, especially on the {\it A}, and to be preceded by
a slight pause. Initialisms, as noted in the context of the discussion
of grounding in \citep{me-cup}, do seem to have distinctive
prosody. 2) The {\it too} of {\it me too}, seemed to be stressed in
most productions, being loud and high in pitch.  Pragmatically, of all
the original phrases the reenactors heard, this was the one with the
most positive feeling.  3) The word {\it summer} also was generally
high in pitch and preceded by a slight pause.  While the word was not
novel in the context, in this phrase it was used in an unusual, novel
sense, to refer to either the first summer half-semester or the second
summer half-semester.  This observation aligns with our secondary
hypothesis.

\section{Reduction and Negative Assessment  in Spanish}

There was tendency for negative assessments in Spanish to be reduced,
as seen in, for example:
\begin{itemize}
    \item {\it lo único malo pues lo que \underline{te digo}, que este año casi *no voy a poder \underline{ir a Chihuahua}.} (001@5:49)

 {\it the only bad thing is what I'm  \underline{telling you}, that this year I'm  *not really going to be able to \underline{go to Chihuahua}.} 

\item {\it\ldots bueno, me dijo, no tengo nada, y no encontre nada en mi mochila y dije  aquí  ya acabo, *ya me saqué \underline{un cero}.} (003@0:46)

 [about taking an exam, and lacking a pencil]        {\it well, he told me, I don't have anything, and I didn't find anything in my backpack and I said, it's all over, *I already got \underline{a zero}.} 

\item {\it A mi *\underline{no} me gusta \underline{el huevo} con tocino.} (012@0:02)

       {\it I *\underline{don't} like \underline{egg} with bacon.} 

\item {\it *\underline{no} me gusta, \underline{ es que} no sabe igual.} (012@0:19)

  {\it {I don't} like it, \underline{ it just} doesn't taste the same.} 

\end{itemize}

As a side note, we there is no reason to think that these various
functions in Spanish are confusable.  Rather, we noticed that each
function may have its own set of characteristic features in addition
to reduction.  While there is a lot of variation, at least sometimes
positive utterances have higher pitch, turn grabs many interleaved
pauses, and downplayed phrases fast rate, low pitch, and lower
intensity.  Further, negative assessments may have, in addition to the
weak tendency In addition to reduction, pitch downslope and a general
slowing over the phrase.

\section{Summary}

The main contributions presented in this technical report, beyond those
reported in the article, are: 
\begin{enumerate} \setlength{\itemsep}{0em}
\item The first publicly released collection
of data annotated for perceived reduction.
 \item The finding that reduction in dialog differs from reduction in
   read speech in its prosodic correlates.
 \item The finding that
   it seems only marginally possible to predict perceived reduction
   from prosodic correlates.
 \item An expanded listing of  functions that may involve reduction. 

\end{enumerate}

\subsubsection*{Acknowledgments}
We thank Divette Marco for the second set of reduction annotations.
We thank Raul O. Gomez and Georgina Bugarini for discussion and for
continuing this work, as reported in the journal article.  We thank
Jose Perez for advice on designing the neural network models.  We
thank Oliver Niebuhr, Visar Berisha, Jonathan Avila, Natasha Warner,
and Rory Turnbull for discussion.  This work was supported in part by
the AI Research Institutes program of the National Science Foundation
and the Institute of Education Sciences, U.S. Department of Education,
through Award \#2229873 -- National AI Institute for Exceptional
Education.

{\small \bibliography{bib}

\begin{figure}[thp]
      \centering
      \includegraphics[scale=0.58, clip, trim=1.6cm 0 0 3.9cm]{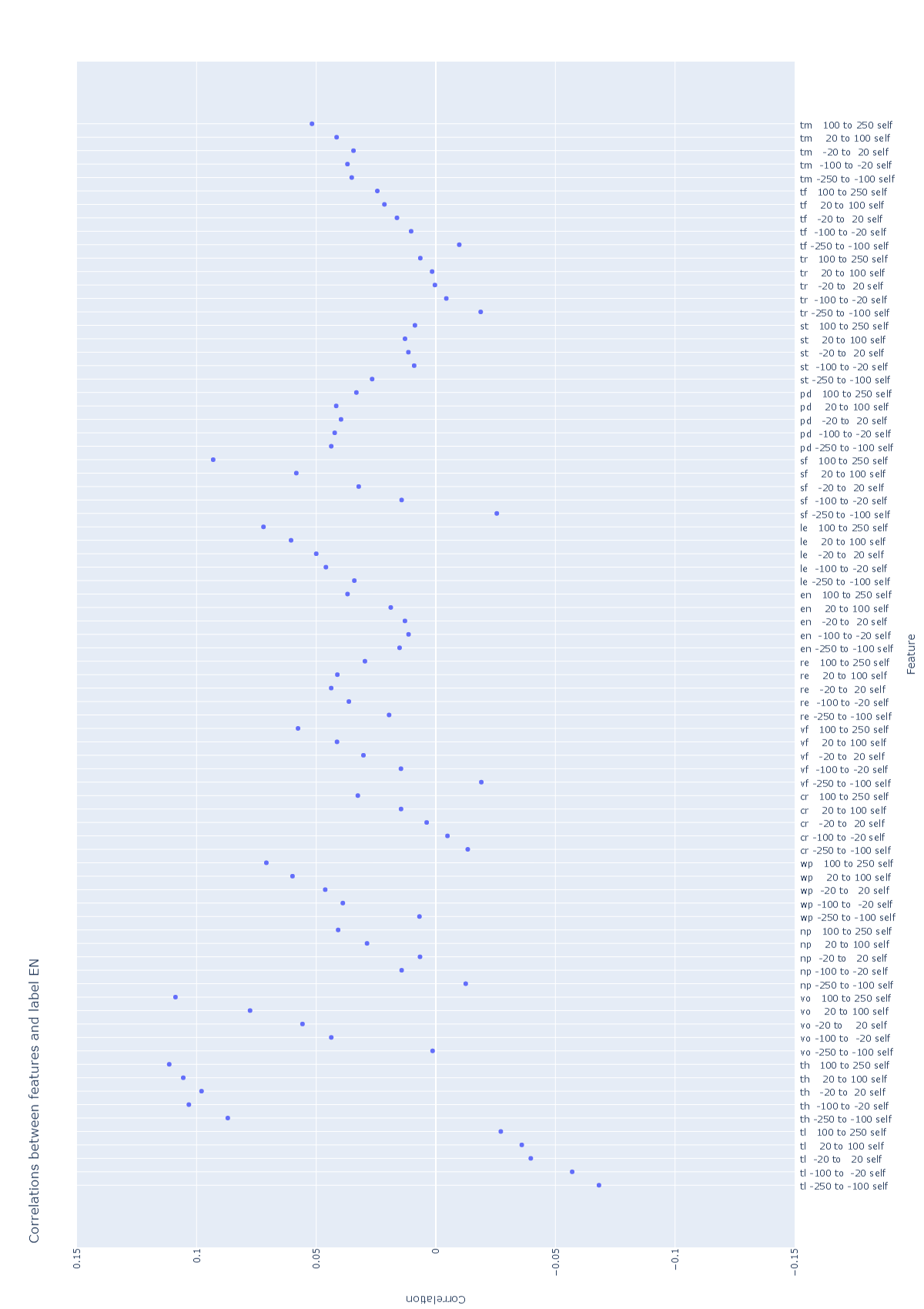}
      \caption{Feature correlations with reduction for English}
      \label{fig:corr-en}
\end{figure}

\begin{figure}[thp]
      \centering
      \includegraphics[scale=0.58, clip, trim=1.6cm 0 0 3.9cm]{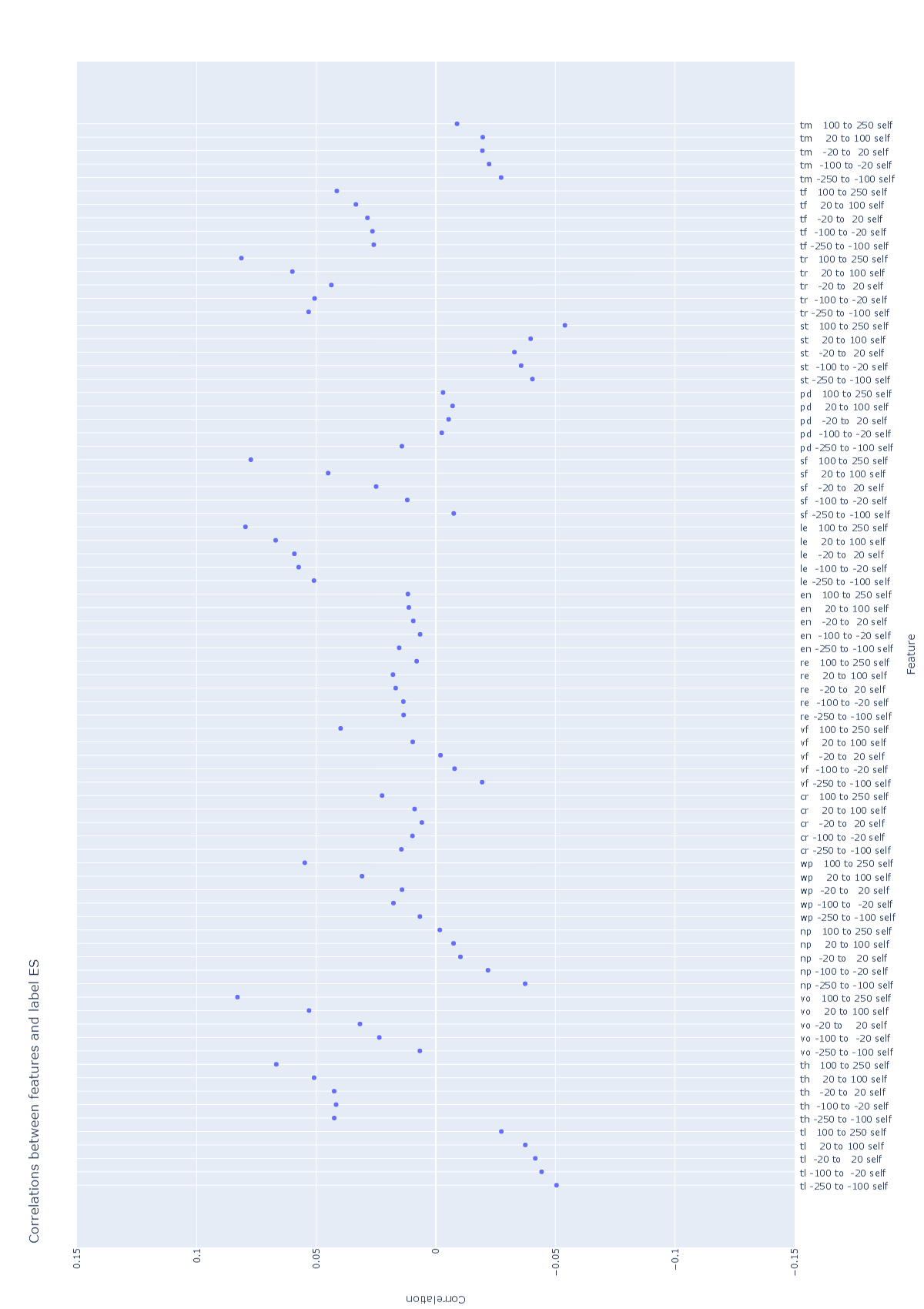}
      \caption{Feature correlations with reduction for Spanish}
      \label{fig:corr-es}
\end{figure}

\end{document}